%% file: bare_jrnl.tex
\pdfoutput=1

\documentclass[journal]{IEEEtran}
\usepackage{algorithm}

\usepackage{amsmath}
\usepackage{amssymb}
\usepackage{xcolor}
\usepackage{algorithmicx}
\usepackage{algcompatible}
\usepackage{algpseudocode}
\usepackage{multirow}
\usepackage{url}

\ifCLASSINFOpdf
  \usepackage[pdftex]{graphicx}
\else
  \usepackage[dvips]{graphicx}
\fi

\begin{document}
\title{Multi-robot Learning-based Informative Path Planning Using Spatio-Temporal Gaussian Process Kalman Filter}

\author{Muqing Cao*$^{1}$, Yunwoo Lee*$^{1}$, Junbin Yuan$^2$, Lorenzo Schenk$^{1}$, and Sebastian Scherer$^1$
\thanks{* Muqing Cao and Yunwoo Lee contributed equally to this work.}
\thanks{Manuscript received: Jun 17, 2026; Accepted October 12, 2026. This paper was recommended for publication by Editor M. Ani Hsieh upon evaluation of the Associate Editor and Reviewers' comments. This work is supported by the Office of Naval Research under Grant N000142512232. Additionally, this work was supported by the DSTA under contract number \#DST000EC124000205}%
\thanks{$^{1}$Authors are with the Robotics Institute, School of Computer Science at Carnegie Mellon University, Pittsburgh, PA, USA}%
\thanks{$^{2}$Author is with the Department of Mechanical Engineering at Carnegie Mellon University, Pittsburgh, PA, USA}%
\thanks{Digital Object Identifier (DOI): see top of this page.}
\thanks{\textcopyright~2026 IEEE. Personal use of this material is permitted. Permission from IEEE must be obtained for all other uses, in any current or future media, including reprinting/republishing this material for advertising or promotional purposes, creating new collective works, for resale or redistribution to servers or lists, or reuse of any copyrighted component of this work in other works.}
}

\markboth{IEEE Robotics and Automation Letters. Preprint Version. Accepted October, 2026}{Cao \MakeLowercase{\textit{et al.}}: Multi-robot Learning-based IPP Using Spatio-Temporal GP Kalman Filter}

\maketitle

\begin{abstract}
Multi-robot informative path planning (IPP) for persistent target monitoring requires robots to reason about spatial uncertainty, temporal evolution, and practical sensing and communication constraints.
Recent learning-based multi-robot IPP methods use Gaussian Processes (GPs) for target uncertainty, but often rely on simplified sensing models and centralized belief updates.
We propose a grid-based spatio-temporal GP--Kalman filtering framework for learning-based multi-robot IPP.
Instead of maintaining one GP per target, we represent anonymous target presence as a single latent field over a discrete workspace grid.
The proposed recursive update assimilates footprint-level camera measurements and supports arbitrary fields of view and range-dependent noise.
A GP-derived temporal process update inflates uncertainty over time to account for moving targets.
For decentralized deployment, each robot maintains its own mapper and exchanges fixed-size belief summaries rather than raw measurements.
Received beliefs are fused using diagonal covariance intersection to remain conservative under unknown inter-robot correlations.
We integrate the mapper with a reinforcement-learning policy for graph-based neighbor selection.
In simulation, the centralized deployment improves target visitation and attains about $20\%$ lower posterior target uncertainty than learning-based and classical auction/coverage baselines, a large share of which stems from the belief representation itself. A real-world two-UAV experiment demonstrates transfer to outdoor multi-robot search over a large field ($>7000\text{m}^2$).
Code is available at \url{https://github.com/caomuqing/multirobot-ipp}.
\end{abstract}


\IEEEpeerreviewmaketitle

\input{intro}

\section{Problem Statement and Overview}
\label{sec:problem}

We study {multi-robot persistent monitoring} in a bounded planar workspace 
$\mathcal{W}\subset\mathbb{R}^2$.
A team of $R$ mobile robots must repeatedly sense the environment to maintain an accurate and up-to-date belief about {dynamic target presence} over space.

Each robot carries a camera sensor with limited field of view (FoV). At time $t_k$, robot $r\in \{ 1,2,\ldots, R\}$ observes a subset of the workspace determined by its pose and the camera FoV.
The camera yaw is aligned with the robot's direction of travel, so the footprint faces the motion direction; it is
not an independent decision variable, and all compared methods in Sec.~\ref{sec:simulation} share this convention.

{The team objective is to maximize the information gain over the map throughout the mission}, i.e., the reduction in global posterior uncertainty over space and time under motion and sensing constraints.

Fig.~\ref{fig:overview} illustrates the proposed decentralized framework: each robot runs onboard target detection, GP--Kalman belief mapping over visible grid cells, periodic belief exchange and fusion, and a learned attention-based policy that selects the next neighboring graph node.

\section{Belief Modeling}
\label{sec:belief}

Each robot represents target presence as a Gaussian belief over a workspace grid and updates it via
(i) a Kalman measurement update under a realistic camera model, (ii) a GP-consistent temporal process update, and,
for scalable decentralized operation, (iii) a diagonal covariance approximation with a conservative belief-fusion rule.
The Kalman--GP formulation itself is prior work: the measurement update follows GP fusion with integral
kernels \cite{jin2022gpf_integral_kernels} and the process-model construction follows Reece and Roberts
\cite{reece2010kfgp}. Our contributions are what is fed into and layered on this machinery: the footprint-level
measurement set with range-dependent noise (Secs.~\ref{subsec:meas-model}--\ref{subsec:meas-update}), the closed-form
Mat\'ern-matched persistence coefficient (Sec.~\ref{subsec:process}), and the diagonal approximation with
kernel-consistent correlations and diagonal-CI fusion
(Secs.~\ref{subsec:diag-approx}--\ref{subsec:ci}) that make the belief compact enough for decentralized exchange.

\subsection{Discrete workspace belief as a latent spatio-temporal field}
We discretize the bounded workspace $\mathcal{W}\subset \mathbb{R}^2$ into a $G\times G$ grid with
$V=G^2$ cells. Let $x_i\in\mathbb{R}^2$ denote the center of cell $i\in\{1,\dots,V\}$.
We model a latent spatio-temporal field $f(x,t)\in\mathbb{R}$ indicating a target-presence probability at location $x$ and time $t$.

At discrete time $t_k$, define the grid state vector
\begin{equation}
\mathbf{f}_k = \big[f(x_1,t_k),\dots,f(x_V,t_k)\big]^\top \in \mathbb{R}^V.
\end{equation}
Each robot maintains a Gaussian belief
\begin{equation}
p(\mathbf{f}_k \mid \mathcal{D}_{1:k}) \;=\; \mathcal{N}(\mathbf{m}_{k|k},\, \mathbf{P}_{k|k}),
\label{eq:belief}
\end{equation}
where $\mathbf{m}_{k|k}\in\mathbb{R}^V$ is the posterior mean map and
$\mathbf{P}_{k|k}\in\mathbb{R}^{V\times V}$ is the posterior covariance.
Here $\mathcal{D}_{1:k}$ denotes the robot's information up to $t_k$: its own measurement history and any beliefs received from teammates under intermittent communication.

In contrast to multi-GP formulations that maintain one GP per target, we maintain a {single} belief map over the workspace, sacrificing target identity. This choice improves scalability for both inference and decentralized belief exchange.
We assume a fixed spatial Gaussian-process prior at each time slice:
\begin{equation}
\mathbf{f}_k \sim \mathcal{N}(\mathbf{m}_0,\mathbf{K}),\qquad
\mathbf{K}_{ij}=k_s(x_i,x_j),
\label{eq:spatial-prior}
\end{equation}
where $\mathbf{m}_0\in\mathbb{R}^V$ is a prior mean map (often uniform/constant) and
$k_s(\cdot,\cdot)$ is a stationary spatial kernel (e.g., squared exponential or Mat\'ern).
We use the integral kernel GP \cite{jin2022gpf_integral_kernels}, where covariance between cell $i$ and $j$ is obtained
by integrating the pointwise kernel over the cell areas.
This makes the GP prior consistent with rasterized map values and reduces discretization artifacts
\cite{jin2022gpf_integral_kernels}.

\subsection{Camera measurement model}
\label{subsec:meas-model}
At time $t_k$, robot $r\in\{1,\dots,R\}$ observes the {visible} grid cells determined by its camera
field-of-view (FoV) and line-of-sight constraints, and assimilates a set $\mathcal{I}_{k}^{(r)}\subset\{1,\dots,V\}$:
all visible cells (as negative observations) if no target is detected, and only the cells containing a detection otherwise.
Let $m_k^{(r)} \triangleq |\mathcal{I}_k^{(r)}|$.

The measurement vector is $\mathbf{z}_{k}^{(r)}\in\mathbb{R}^{m_k^{(r)}}$ with entries
$z_{k}^{(r)}(i)\in\{0,\bar z\}$ ($\bar z>0$ fixed) indicating binary detection of target presence in cell $i\in\mathcal{I}_k^{(r)}$.
To enable Kalman-style fusion, we adopt a standard linear-Gaussian approximation:
\begin{equation}
\mathbf{z}_{k}^{(r)} = \mathbf{H}_{k}^{(r)}\mathbf{f}_k + \mathbf{v}_{k}^{(r)},\qquad
\mathbf{v}_{k}^{(r)}\sim \mathcal{N}(\mathbf{0},\mathbf{R}_{k}^{(r)}),
\label{eq:lin-gauss}
\end{equation}
where $\mathbf{H}_k^{(r)}\in\{0,1\}^{m_k^{(r)}\times V}$ is a selection matrix whose rows each pick one grid index in $\mathcal{I}_k^{(r)}$.
Relaxing the binary detections to additive Gaussian noise is standard in integral-kernel GP fusion
\cite{jin2022gpf_integral_kernels}. It is an approximation rather than a bound: at short range the noise variance
of \eqref{eq:noise-model} can be smaller than that of the underlying Bernoulli detection, so the update may be locally
overconfident, which is one reason we also evaluate with ground-truth visit counts (Sec.~\ref{subsec:sim-metrics}).

\paragraph{Distance/geometry-dependent noise.}
Sensing degradation is encoded through a diagonal noise covariance
\begin{equation}
\mathbf{R}_{k}^{(r)}=\mathrm{diag}\!\left(\sigma_{k}^{2}(i)\right)_{i\in\mathcal{I}_k^{(r)}},\qquad
\sigma_{k}(i)=\sigma_0\cdot g\!\left(d_k^{(r)}(i),\theta_k^{(r)}(i)\right),
\label{eq:noise-model}
\end{equation}
where $d_k^{(r)}(i)$ is the range from robot $r$ to cell $i$, $\theta_k^{(r)}(i)$ may include viewing geometry, and $g(\cdot)$ is increasing in distance.

\subsection{Sequential measurement update}
\label{subsec:meas-update}
Let $(\mathbf{m}_{k|k-1},\mathbf{P}_{k|k-1})$ denote the predicted (prior) belief at time $t_k$ before incorporating
robot $r$'s measurement.
Assimilating \eqref{eq:lin-gauss} via the Kalman update is equivalent to GP fusion over the discretized state
\cite{jin2022gpf_integral_kernels}:
\begin{align}
\mathbf{S}_{k}^{(r)} &\triangleq \mathbf{H}_{k}^{(r)}\mathbf{P}_{k|k-1}\mathbf{H}_{k}^{(r)\top} + \mathbf{R}_{k}^{(r)},
\label{eq:S}\\
\boldsymbol{\Gamma}_{k}^{(r)} &\triangleq \mathbf{P}_{k|k-1}\mathbf{H}_{k}^{(r)\top}\big(\mathbf{S}_{k}^{(r)}\big)^{-1},
\label{eq:gain}\\
\mathbf{m}_{k|k}^{(r)} &= \mathbf{m}_{k|k-1} + \boldsymbol{\Gamma}_{k}^{(r)}\big(\mathbf{z}_{k}^{(r)}-\mathbf{H}_{k}^{(r)}\mathbf{m}_{k|k-1}\big),
\label{eq:m-update}\\
\mathbf{P}_{k|k}^{(r)} &= \mathbf{P}_{k|k-1} - \boldsymbol{\Gamma}_{k}^{(r)}\mathbf{S}_{k}^{(r)}\boldsymbol{\Gamma}_{k}^{(r)\top}.
\label{eq:P-update}
\end{align}
Here $\mathbf{S}_{k}^{(r)}$ is the {innovation covariance} and $\boldsymbol{\Gamma}_{k}^{(r)}$ is the {Kalman gain},
which propagates each measurement to all grid cells through the spatial covariance.
Because $m_k^{(r)}$ is small (the FoV is small), inverting $\mathbf{S}_k^{(r)}$ is cheap and the update
applies sequentially as new measurements arrive \cite{jin2022gpf_integral_kernels}.

\subsection{GP-consistent Kalman process update}
\label{subsec:process}

Between measurement updates targets move and sensing becomes stale, so a {process update} must propagate the mean
belief forward and inflate uncertainty in a principled way; following \cite{reece2010kfgp}, we derive it from a
spatio-temporal GP prior.
For the discretized grid state $\mathbf{f}_k=[f(x_1,t_k),\dots,f(x_V,t_k)]^\top$, define the
same-time covariances $\mathbf{K}_{k-1}\triangleq\mathrm{Cov}(\mathbf{f}_{k-1},\mathbf{f}_{k-1})$,
$\mathbf{K}_{k}\triangleq\mathrm{Cov}(\mathbf{f}_{k},\mathbf{f}_{k})$ and the cross-time covariance
$\mathbf{C}_{k-1}\triangleq\mathrm{Cov}(\mathbf{f}_{k},\mathbf{f}_{k-1})$, all in $\mathbb{R}^{V\times V}$.
Consider the GP joint prior over two consecutive time slices:
\begin{equation}
\begin{bmatrix}
\mathbf{f}_k\\
\mathbf{f}_{k-1}
\end{bmatrix}
\sim
\mathcal{N}\!\left(
\begin{bmatrix}
\mathbf{m}_{0,k}\\
\mathbf{m}_{0,k-1}
\end{bmatrix},
\begin{bmatrix}
\mathbf{K}_k & \mathbf{C}_{k-1}\\
\mathbf{C}_{k-1}^\top & \mathbf{K}_{k-1}
\end{bmatrix}
\right),
\label{eq:joint}
\end{equation}
where $\mathbf{m}_{0,k}$ denotes the GP prior mean at time $t_k$ (in our experiments, we use a time-invariant prior mean $\mathbf{m}_{0,k}=\mathbf{m}_0$).

Reece and Roberts \cite{reece2010kfgp} show that this conditional distribution
$p(\mathbf{f}_k\mid \mathbf{f}_{k-1})$ can be written as a linear-Gaussian Markov transition (if the block covariance in \eqref{eq:joint} is positive semidefinite):
\begin{equation}
\mathbf{f}_k
=
\mathbf{m}_{0,k}
+
\mathbf{G}_{k-1}\big(\mathbf{f}_{k-1}-\mathbf{m}_{0,k-1}\big)
+
\mathbf{w}_k,
\;
\mathbf{w}_k\sim\mathcal{N}(\mathbf{0},\mathbf{Q}_k),
\label{eq:process}
\end{equation}
\begin{equation}
\mathbf{G}_{k-1}=\mathbf{C}_{k-1}\mathbf{K}_{k-1}^{-1},
\;
\mathbf{Q}_k=\mathbf{K}_k-\mathbf{C}_{k-1}\mathbf{K}_{k-1}^{-1}\mathbf{C}_{k-1}^\top.
\label{eq:GQ}
\end{equation}
Here $\mathbf{G}_{k-1}$ acts as a persistence operator and $\mathbf{Q}_k$ injects uncertainty; the Kalman prediction step follows from \eqref{eq:process}.

To connect this process model with a temporal GP kernel, consider a separable spatio-temporal covariance
\begin{equation}
k\big((x,t),(x',t')\big)=k_s(x,x')\,k_t(t,t'),
\label{eq:separable}
\end{equation}
where $k_s$ and $k_t$ are scalar spatial and temporal kernels. If $k_t$ is stationary, we write
$k_t(t,t')=k_t(\tau)$ with $\tau=|t-t'|$.
Let $\Delta t = t_k-t_{k-1}$.
Then the grid-level covariances induced by \eqref{eq:separable} satisfy 

\begin{equation}
\mathbf{K}_{k-1}=\mathbf{K}_k=\mathbf{K},\,
\mathbf{C}_{k-1}=\alpha(\Delta t)\mathbf{K},\,
\alpha(\Delta t)\triangleq \frac{k_t(\Delta t)}{k_t(0)}.
\label{eq:Csep}
\end{equation}
Substituting \eqref{eq:Csep} into \eqref{eq:GQ} yields
\begin{equation}
\mathbf{G}_{k-1}=\alpha\,\mathbf{I}_V,\qquad
\mathbf{Q}_k=(1-\alpha^2)\mathbf{K},
\label{eq:alpha-def}
\end{equation}
where $\mathbf{I}_V$ is the identity, so $\alpha$ quantifies how strongly the field persists from $t_{k-1}$ to $t_k$.

With $\mathbf{m}_{0,k}=\mathbf{m}_0$ and $\mathbf{K}$ fixed, the prediction step becomes
\begin{align}
\mathbf{m}_{k|k-1} &= \mathbf{m}_0 + \alpha\big(\mathbf{m}_{k-1|k-1}-\mathbf{m}_0\big), \label{eq:alpha-mean}\\
\mathbf{P}_{k|k-1} &= \alpha^2\mathbf{P}_{k-1|k-1} + (1-\alpha^2)\mathbf{K}. \label{eq:alpha-cov}
\end{align}
For a stationary Mat\'ern-$\tfrac{3}{2}$ temporal covariance kernel with
length-scale $\ell_t>0$, the normalized temporal correlation is
\begin{equation}
\alpha(\Delta t)
\triangleq
\frac{k_t(\Delta t)}{k_t(0)}
=
\left(1+\sqrt{3}\frac{\Delta t}{\ell_t}\right)
\exp\!\left(-\sqrt{3}\frac{\Delta t}{\ell_t}\right).
\label{eq:alpha-matern32}
\end{equation}
Thus, $\alpha(\Delta t)$ directly gives the one-step persistence coefficient used in the Kalman process update.
Being first order, the update matches the Mat\'ern-$\tfrac{3}{2}$ correlation only at the single lag $\Delta t$;
over several steps, correlations decay exponentially.

\subsection{Diagonal-covariance approximation with fixed spatial correlations}
\label{subsec:diag-approx}

Maintaining the full covariance $\mathbf{P}_{k|k}$ is $O(V^2)$ in memory and communication.
We therefore store only the marginal variances
\begin{equation}
\mathbf{s}_{k|k}\triangleq \mathrm{diag}(\mathbf{P}_{k|k})\in\mathbb{R}^V,
\end{equation}
and reconstruct an approximate covariance from the updated marginal variances and a fixed kernel-derived spatial
correlation pattern, leveraging the standard variance--correlation decomposition
$\mathbf{\Sigma}=\mathbf{D}^{1/2}\mathbf{R}\mathbf{D}^{1/2}$ \cite{anderson2003multivariate}.
Define the kernel-induced correlation matrix
\begin{equation}
\mathbf{D}_K\triangleq \mathrm{diag}\!\big(\mathrm{diag}(\mathbf{K})\big),\qquad
\mathbf{R}_K\triangleq \mathbf{D}_K^{-1/2}\mathbf{K}\mathbf{D}_K^{-1/2},
\label{eq:RK}
\end{equation}
where $\mathbf{D}_K$ holds the prior variances and $\mathbf{R}_K$ is a correlation matrix.
Given $\mathbf{s}_{k|k}$, define $\mathbf{D}_{k|k}\triangleq \mathrm{diag}(\mathbf{s}_{k|k})$ and approximate
\begin{equation}
\widetilde{\mathbf{P}}_{k|k}
\triangleq \mathbf{D}_{k|k}^{1/2}\,\mathbf{R}_K\,\mathbf{D}_{k|k}^{1/2}.
\label{eq:Ptilde}
\end{equation}
The approximation is exact if the posterior correlation equals $\mathbf{R}_K$; otherwise it still preserves the
marginals $\mathrm{diag}(\widetilde{\mathbf{P}}_{k|k})=\mathbf{s}_{k|k}$ exactly together with a kernel-consistent
correlation template, enabling scalable mapping and communication  (its fidelity and closed-loop cost are
quantified in Sec.~\ref{subsec:ablations}).

\subsection{Decentralized belief fusion via diagonal CI}
\label{subsec:ci}
Robots communicate intermittently, so cross-correlations between robot beliefs are unknown and naive fusion can
double-count information. Covariance Intersection (CI) provides a conservative fusion rule that remains consistent
under unknown cross-correlation \cite{julier1997ci,uhlmann2001ci}.

Given two beliefs $\mathcal{N}(\mathbf{m}_a,\mathbf{P}_a)$ and
$\mathcal{N}(\mathbf{m}_b,\mathbf{P}_b)$, CI forms
\begin{align}
\mathbf{P}_c^{-1} &= \omega \mathbf{P}_a^{-1} + (1-\omega)\mathbf{P}_b^{-1},
\qquad \omega\in[0,1], \label{eq:ciP}\\
\mathbf{m}_c &= \mathbf{P}_c
\left(\omega \mathbf{P}_a^{-1}\mathbf{m}_a
+ (1-\omega)\mathbf{P}_b^{-1}\mathbf{m}_b\right).
\label{eq:cim}
\end{align}

Here subscript $c$ denotes the fused belief and $\omega\in[0,1]$ is a scalar {fusion weight} trading off the two
sources; it can be optimized by a 1D line search over $\mathrm{tr}(\mathbf{P}_c)$, or fixed (e.g., $\omega=0.5$)
for simplicity \cite{julier1997ci,uhlmann2001ci}. Sec.~\ref{subsec:ablations} shows the two approaches perform similarly
in our setting, although a scalar $\omega$ also down-weights a sender's information in cells only it has observed.

Under diagonal storage (using $\mathbf{s}=\mathrm{diag}(\mathbf{P})$), \eqref{eq:ciP}--\eqref{eq:cim} reduce to
element-wise fusion per grid cell.

\section{RL-based Planning}
\label{sec:rl}

We cast multi-robot planning as a Decentralized Partially Observable Markov Decision Process
(Dec-POMDP). Following the design of \cite{zhang2025compass}, each robot executes a shared policy that maps local spatio-temporal belief features on a graph
to a distribution over one-hop neighbor actions.

Free space is discretized into a navigation graph
$\mathcal{G}=(\mathcal{V},\mathcal{E})$ with $|\mathcal{V}|=K$ waypoints and $k$-NN edges in collision-free space.
At decision step $k$, robot $r$ occupies a node $v^{(r)}_k\in\mathcal{V}$ and chooses its next node among
one-hop neighbors $\mathcal{N}(v^{(r)}_k)\triangleq\{u\mid (v^{(r)}_k,u)\in\mathcal{E}\}$.
The motion model is $v^{(r)}_{k+1} = a^{(r)}_k$ with $a^{(r)}_k \in \mathcal{N}(v^{(r)}_k)$.

Each robot receives an observation $o^{(r)}_k\in\mathcal{O}^m$ derived from its local belief and neighborhood features
(defined below), and the team receives a shared reward $r_k=r(s_k,\mathbf{a}_k)$.
We optimize the expected discounted return $\mathbb{E}[\sum_{k=0}^{T-1}\gamma^k r_k]$ for the mission duration $T$.

\subsection{Belief-to-graph feature construction}
Planning operates on the waypoint graph $\mathcal{G}=(\mathcal{V},\mathcal{E})$, while inference maintains a grid belief
over $\mathbf{f}_k$ (Sec.~\ref{sec:belief}).
To connect the two, each graph node $v\in\mathcal{V}$ is assigned a corresponding grid index $\iota(v)\in\{1,\dots,V\}$
(e.g., nearest grid-cell center), and we define a node-level belief feature from the map posterior,
together with one-step predictions under the GP-consistent process model
\eqref{eq:alpha-mean}--\eqref{eq:alpha-cov}:
\begin{align}
\phi_k(v) &\triangleq \big[ m_{k|k}(\iota(v)), s_{k|k}(\iota(v)) \big],\nonumber\\
\phi^{\mathrm{pred}}_k(v) &\triangleq \big[ m_{k+\delta|k}(\iota(v)), s_{k+\delta|k}(\iota(v)) \big],
\end{align}
where $\delta$ corresponds to the next decision time.
We further append (i) a binary presence flag $p_k(v)\in\{0,1\}$ indicating whether any robot currently occupies $v$,
and (ii) geometric features including 2D coordinates, coord($v$), and positional encodings on the graph, PE($v$).
The final node feature is
$x_k(v) \triangleq \big[\phi_k(v),\phi^{\mathrm{pred}}_k(v),p_k(v),\mathrm{coord}(v),\mathrm{PE}(v)\big]$.
Compared to COMPASS-style per-target features, our node representation contains {only one} mean/variance pair
because we maintain a single workspace belief field rather than one GP per target identity.

\subsection{Observation and action spaces}
At decision step $k$, robot $r$ observes the feature history of its current node and one-hop neighbors,
$o^{(r)}_k \triangleq \{x_{k-\ell}(v)\,|\,v\in\{v^{(r)}_k\}\cup\mathcal{N}(v^{(r)}_k),\,\ell=0,\dots,L_{\mathrm{hist}}-1\}$
with temporal history length $L_{\mathrm{hist}}$, and its discrete action
$a^{(r)}_k \in \mathcal{N}(v^{(r)}_k)$ selects the next neighbor node to visit.

We employ a shared spatio-temporal attention backbone similar in spirit to COMPASS \cite{zhang2025compass}:
a {temporal encoder} aggregates each node's feature history, a {spatial encoder} attends over graph nodes,
and an {actor head} scores the candidate neighbors $\mathcal{N}(v^{(r)}_k)$ to produce
$\pi_\theta(a^{(r)}_k\mid o^{(r)}_k)$ (a {critic head} estimates values).
All robots share parameters $\theta$, enabling transfer across team sizes.

\subsection{Reward design}
Following \cite{zhang2025compass}, the step reward encourages uncertainty reduction at the targets while penalizing
redundant sensing and travel:
\begin{equation}
r_k \;=\; \lambda_{\mathrm{info}}\,\mathrm{IG}_k
\;-\;\lambda_{\mathrm{red}}\,\mathrm{RP}_k
\;-\;\lambda_{\mathrm{path}}\,\mathrm{PC}_k.
\end{equation}
{Information gain} sums, over targets $n$, the reduction $\Delta\sigma_{n,k}\geq 0$ of the posterior standard
deviation at the cell containing target $n$ (on the reference belief of Sec.~\ref{subsec:sim-metrics} when
decentralized), clipped at $0.2$ and shared among the $c_{n,k}$ robots observing it:
$\mathrm{IG}_k=\sum_{n}\min(\Delta\sigma_{n,k},0.2)/\max(c_{n,k},1)$.
{Redundancy penalty}
$\mathrm{RP}_k \triangleq \frac{1}{R}\sum_{n} c_{n,k}^2\,(1-\bar u_{n,k})$
discourages repeated observation of already-certain targets, with $\bar u_{n,k}$ the running mean uncertainty of
target $n$; {path cost}
$\mathrm{PC}_k \triangleq \frac{1}{R}\sum_{r} d_{\mathcal{G}}(v^{(r)}_k, v^{(r)}_{k+1})^{1.5}$
penalizes travel via edge lengths $d_{\mathcal{G}}$ on $\mathcal{G}$.
We use fixed weights $(\lambda_{\mathrm{info}},\lambda_{\mathrm{red}},\lambda_{\mathrm{path}})=(2,0.4,0.1)$; target
positions enter only this training reward, never the policy's observations.

We train with centralized training and decentralized execution (CTDE)---a centralized critic conditions on joint
information while the actor uses only $o^{(r)}_k$ at test time---and optimize the shared policy with Proximal Policy
Optimization (PPO).

\input{simulation_benchmark}

\input{experiment}

\section{Conclusion}\label{sec: conclu}
This paper presented a practical learning-based multi-robot IPP framework that combines a grid-based spatio-temporal GP--Kalman mapper, realistic footprint-level sensing updates, and lightweight decentralized belief fusion to improve persistent monitoring under imperfect sensing and communication. 
In simulation, the centralized deployment improves target visitation and reduces posterior target uncertainty over learning-based and heuristic baselines, in large part through the representation, although this uncertainty is self-reported; the decentralized deployment exchanges only belief summaries at a cost in performance, especially on the robots' own beliefs, and a two-UAV field run demonstrates transfer to hardware. Future work will consider controlled field comparisons, lossy communication, occlusion models, and larger-scale deployments.

\ifCLASSOPTIONcaptionsoff
  \newpage
\fi

\bibliographystyle{IEEEtran}

\bibliography{ref}

\end{document}

%% file: intro.tex
\section{Introduction}
\label{sec:intro}
Multi-robot {informative path planning} (IPP) seeks coordinated motion policies that collect measurements to
reduce uncertainty about an environment of interest.
A canonical instance is {persistent monitoring} for target search and tracking: a team of robots must
repeatedly sense a workspace under motion and sensing budgets while targets move, updating the target-presence
belief online from imperfect sensing and coordinating over bandwidth-limited communication.

Gaussian Processes (GPs) are attractive belief models for monitoring because they provide a non-parametric prior over
functions with calibrated posterior uncertainty \cite{rasmussen2006gpml}.
Recent learning-based IPP methods, such as CAtNIPP \cite{cao2023catnipp} and COMPASS \cite{zhang2025compass}, combine GP
belief estimation with attention-based policy networks and enable scalable decision-making on graph abstractions.
However, existing learning-based GP-based {multi-robot} IPP methods still rely on simplified sensing models and strict information-sharing assumptions that hinder real-world deployment. 

First, sensing is commonly abstracted as a fixed-radius {circular} region and beliefs are updated using coarse binary detections.
Specifically, learning-based methods allow an observing robot to obtain at most one binary observation per target and decision step \cite{zhang2025compass}.
This abstraction yields an efficient learning interface but does not capture {footprint-level} evidence within the
camera field-of-view (FoV), nor range-dependent sensing quality.
Moreover, maintaining one GP per target assumes the number of targets is known a priori and complicates deployment when
target identities are ambiguous.
Second, to maintain a consistent team belief, GP updates must aggregate information across the team.
COMPASS synchronizes beliefs by aggregating the raw measurements from all agents after each measurement
cycle \cite{zhang2025compass}, which can be challenging under bandwidth-limited and intermittent communication. 

To address these issues, this letter proposes a discrete, grid-based belief representation and a recursive
{GP--Kalman} update that bridges GP priors and sequential Bayesian fusion.
We represent target presence as a latent spatio-temporal field on a grid and maintain a Gaussian belief over
grid values.
Following GP fusion with integral kernels \cite{jin2022gpf_integral_kernels}, the measurement update assimilates
evidence from the cells of the sensor footprint rather than a single per-target detection, and supports distance-dependent sensing noise.
To model temporal evolution, we adopt a Kalman process model derived from the joint GP prior across time
\cite{reece2010kfgp}.
For scalable multi-robot deployment, each robot runs its own mapper and exchanges only compact belief summaries; we fuse
received beliefs using a lightweight rule based on {diagonal} covariance intersection (CI)
\cite{julier1997ci,uhlmann2001ci} to remain conservative under unknown inter-robot cross-correlations.

Our main contributions are:
\begin{itemize}
    \item \textbf{Grid-based spatio-temporal GP--Kalman mapper with realistic sensing.}
    We formulate a discrete GP belief over a workspace grid and derive a Kalman measurement update that assimilates
    {footprint-level} camera measurements (with range-dependent noise), together with a
    GP-consistent process update for temporal evolution.
    \item \textbf{Lightweight decentralized belief fusion under limited communication bandwidth.}
    We introduce a diagonal covariance approximation with kernel-consistent fixed correlations and develop an efficient
    diagonal-CI fusion rule that is computationally light and robust to unknown inter-robot correlations. 
    \item \textbf{End-to-end multi-robot IPP integration and evaluation.}
    We integrate the proposed mapper and fusion into a multi-agent reinforcement learning framework
    for graph-based planning, demonstrating improved uncertainty reduction and target coverage over COMPASS and heuristic baselines in simulation.
    \item \textbf{Real-world Experiment.} We conduct real-world persistent monitoring tasks using two drones over a large field ($>7000\text{m}^2$).
\end{itemize}

\begin{figure*}[ht]
    \centering
    \includegraphics[width=0.8\linewidth]{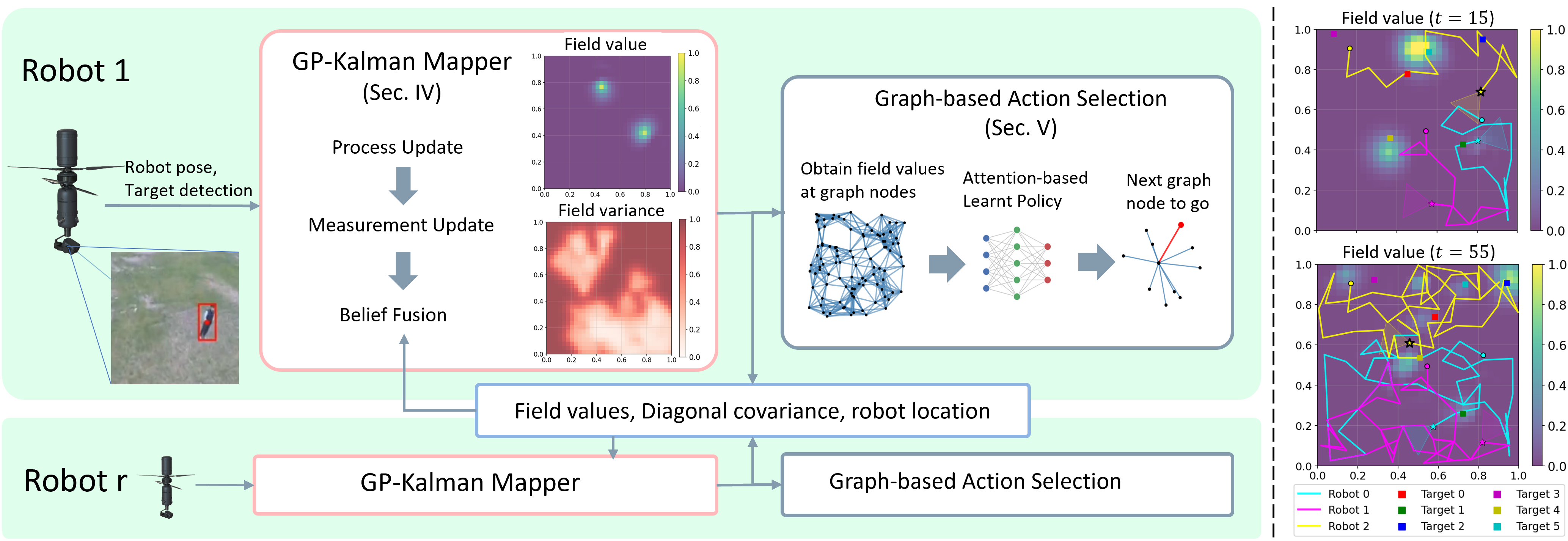}
    \caption{Left: overview of the proposed framework. Right: Visualization of the field mapping result of a simulation run with three robots and six targets. 
    }
    \label{fig:overview}
\end{figure*}

\section{Related Works}
\label{sec:related}

IPP studies how robots move to acquire measurements that maximize an information objective under motion, sensing,
and resource constraints; in persistent monitoring, informative locations must also be {revisited} as the
environment or targets evolve.

A large body of work views environmental monitoring as {active regression} of a continuous latent field, often modeled
as a Gaussian process (GP) to capture spatial correlation \cite{rasmussen2006gpml}.
Classic approaches optimize posterior entropy or mutual information along trajectories
\cite{binney2010informative,hitz2014focused,hollinger2014sampling}.
More recent formulations incorporate localization uncertainty when
planning informative trajectories \cite{popovic2020active_fieldmapping}, or consider online replanning \cite{moon2026ia}.

Scaling IPP to robot teams introduces the need to coordinate which areas are sensed and when.
A common modeling route is to cast multi-robot IPP as a (team) orienteering / traveling-salesman problem, where
reward encodes information gain and tour length encodes resource constraints \cite{gunawan2016orienteering,popovic2020terrain}.
Singh {et al.} exploit the submodularity of common information objectives for efficient multi-robot IPP with
approximation guarantees \cite{singh2007emip,singh2009efficient_sensing}, later extended to non-myopic adaptive
replanning \cite{singh2009nonmyopic}; sparse-optimization formulations scale GP active-regression IPP to
continuous trajectories and robot teams \cite{dutta2025sparseopt,jakkala2024mripp_sparsegp}.

Learning-based IPP has recently emerged as an efficient alternative to hand-crafted heuristics, especially for large-scale
coordination on graphs: CAtNIPP \cite{cao2023catnipp} learns a reactive attention-based policy for adaptive IPP,
STAMP \cite{wang2023stamp} adds temporal attention for single-robot persistent monitoring of mobile targets,
MA-G-PPO \cite{chen2021magppo} and GALOPP \cite{mishra2024galopp} extend multi-agent reinforcement learning (MARL) to
persistent monitoring with sensing and communication constraints, and COMPASS \cite{zhang2025compass} proposes a
cooperative MARL framework with a spatio-temporal attention network.
While these approaches show strong coordination in simulation, most still rely on simplified range-based sensing
and assume frequent information sharing to maintain consistent beliefs \cite{zhang2025compass}; real-world multi-robot deployments
with realistic sensing footprints and limited communication remain comparatively underexplored.

%% file: simulation_benchmark.tex
\section{Simulation}
\label{sec:simulation}

\subsection{Experimental setup}
\label{subsec:sim-setup}
All methods are trained and evaluated in a normalized workspace $\mathcal{W}=[0,1]^2$ containing $N$ mobile targets.
Each robot flies at a fixed normalized altitude of $0.03$ and carries a downward-looking
camera pitched at $55^\circ$ with a $40^\circ$ horizontal and $45^\circ$ vertical
field of view; the resulting non-circular sensor footprint, distance-dependent
measurement noise $\sigma_k(i)=\sigma_0\,(1+\beta\,d_k^{(r)}(i))$ with
$\sigma_0=0.1$ and $\beta=50$ (cf.\ \eqref{eq:noise-model}).
The belief is maintained on a $G\times G=32\times 32$ grid
($V=1024$ cells), with spatial length-scale $0.06$ and a Mat\'ern-$\tfrac32$
temporal decay whose length-scale $\ell_t$ is set to mirror the COMPASS
spatio-temporal GP \cite{zhang2025compass}.

The navigation graph $\mathcal{G}$ is generated by uniformly sampling $200$ nodes
in free space and connecting each to its $k=10$ nearest neighbors.
Robots move one hop per decision step under a per-episode travel budget of $30$
The policy is trained with PPO (learning rate $10^{-4}$, $\gamma=0.99$,
$16$ parallel rollout workers, embedding dimension $128$) under CTDE with randomization of
graph size, budget, and target count, using one training seed per policy; every policy, including the
retrained COMPASS, was trained for more than $5$k episodes and stopped manually once it had stabilized.
We evaluate on team sizes $R\in\{2,3,5\}$ with $N=6$ targets moving at speed
$1/20$, and report metrics averaged over $32$ evaluation episodes per
configuration with target trajectories matched across methods; figures report the mean
with $95\%$ confidence intervals; one network serves every team size.
Unless stated otherwise, all learned methods (both PROPOSED variants and
COMPASS) use the non-collision rule of Sec.~\ref{subsec:nc-ablation}, as deployed
on hardware and matching the classical baselines, which deconflict by
construction; Table~\ref{tab:nc} ablates this choice.

\subsection{Compared methods}
\label{subsec:sim-methods}
\textbf{PROPOSED-Centralized} and \textbf{PROPOSED-Decentralized} are the two
deployments of our approach.
In the centralized variant, a single mapper assimilates the measurements of all
robots to maintain map belief for the shared policy, analogous to the centralized GP update in COMPASS.
In the decentralized variant, each robot maintains its own diagonal GP-Kalman mapper
(Sec.~\ref{subsec:diag-approx}), performs local process/measurement updates from
its own observations, and exchanges its belief (mean and diagonal variance) with teammates every
$3$ steps, fusing them with the conservative diagonal covariance-intersection
rule of Sec.~\ref{subsec:ci} ($\omega=0.5$). No raw measurements are broadcast in
the decentralized case.
Each packet is a fixed $16$\,KB (mean and variance of the $1024$ cells in
double precision), more than a step's raw measurements (about $14$ cells,
${<}0.2$\,KB per robot); its benefit is being self-contained, so a lost or late
packet is superseded by the next without message histories or replay.
Coordination degrades gracefully with sparser exchanges
(Sec.~\ref{subsec:ablations}); simulated links are lossless, zero-latency, and
all-to-all.

\textbf{COMPASS} \cite{zhang2025compass} is the original learning-based
multi-robot informative path planner, which maintains one Gaussian process per
target and a centralized GP update. We adapt its sensing model to the
non-circular camera footprint used here (the released implementation supports
only a circular footprint) so that all methods observe identical measurements,
and retrain it under this sensing model with its original, unmodified reward.
The adaptation changes only the detection test (from a fixed-radius circle
to the heading-dependent footprint of Sec.~\ref{subsec:meas-model})
and adds the same range-dependent noise constants used by our mapper; the
per-target GP belief, network, action space, and training pipeline are
unchanged. 
At evaluation time, COMPASS uses the
same non-collision rule as the PROPOSED variants.

\textbf{AUCTION} is a classical baseline: at each reassignment, candidate nodes
are the most uncertain graph nodes (top $10R$ by per-node posterior variance),
and robots bid on them with a utility that rewards uncertainty and penalizes
travel distance ($u=\text{unc}-0.5\,d$). A greedy conflict-free assignment gives
each robot a persistent target, and the robot follows the true shortest graph
path (BFS over the $k$-NN adjacency) toward it, re-auctioning only upon arrival.

\textbf{COVERAGE}: a tour is formed over a random
sample of waypoints and robots are spread evenly along it, each advancing to the
next tour node via BFS routing on the full graph. It seeks balanced spatial
coverage without using the belief.

\subsection{Metrics}
\label{subsec:sim-metrics}
\textbf{Average uncertainty} (lower is better) is the posterior \emph{standard deviation}
evaluated at the grid cell containing each target, averaged over the $N$ targets
and over the episode:
$\frac{1}{T}\sum_{k=1}^{T}\frac{1}{N}\sum_{n=1}^{N}\sqrt{s_{k|k}(\iota(x^{\mathrm{tgt}}_{n,k}))}$,
where $\iota(x^{\mathrm{tgt}}_{n,k})$ is the index of the grid cell nearest the
$n$-th target at time $t_k$.

For both PROPOSED deployments, this metric is evaluated on a \emph{reference
belief}: a GP--Kalman mapper that assimilates the measurements of all robots
(the policy's own belief when centralized); for the decentralized deployment we
also report it on the robots' own fused beliefs, averaged over robots.
For COMPASS, the metric is
evaluated on its native per-target posterior, probed at the same location and
with matched kernel variance ($1.0$) and measurement-noise constants
($\sigma_0$, $\beta$) as the PROPOSED, so posterior standard deviations are on a common scale.
In both, the posterior variance is independent of the measured values, so the
metric is self-reported, does not measure accuracy, and can favour an
overconfident belief; the visit counts below are measured against ground
truth independently of the belief.
\textbf{Minimum visit count} (higher is better) is the number of observations received by the
{least-observed} target over an episode, capturing worst-case coverage fairness.
\textbf{Average visit count} (higher is better) is the mean number of observations per target
over an episode, reflecting overall sensing throughput on the targets.

\subsection{Results and analysis}
\label{subsec:sim-results}
\begin{figure*}[t]
\centering
\includegraphics[width=0.95\textwidth]{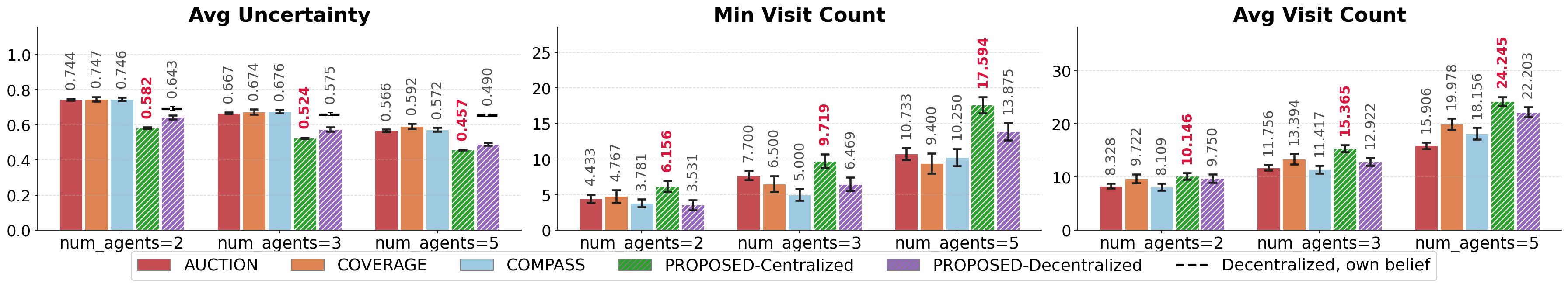}
\caption{Benchmark across team sizes ($R=2,3,5$, $N=6$ targets); best value per group highlighted.
Bars: means over $32$ episodes with $95\%$ CIs; all learned methods use the
non-collision rule (Sec.~\ref{subsec:nc-ablation}); dashes: decentralized robots' own fused beliefs.}
\label{fig:benchmark}
\end{figure*}

Fig.~\ref{fig:overview}(right) shows paths generated by PROPOSED-Decentralized with three robots and six targets; Fig.~\ref{fig:benchmark} reports the three metrics against team size.

\paragraph{Effectiveness of the grid-based GP representation}
On the reference belief, both PROPOSED variants attain markedly lower average uncertainty than COMPASS
and the classical baselines at every team size (e.g.\ at $R=3$,
$0.52$--$0.58$ versus $0.68$ for COMPASS and $0.67$ for AUCTION/COVERAGE),
with the centralized variant about $20\%$ below the strongest baseline. Because COMPASS, AUCTION, COVERAGE and our methods share the
identical sensing model, this gap stems from the representation
and the policy it induces (Sec.~\ref{subsec:ablations} separates the two): the grid belief
assimilates footprint-level evidence, whereas the per-target GPs of COMPASS
admit at most one binary detection per target and step. The PROPOSED
policies also pair this low uncertainty with high average visit counts (e.g.\
$24.2$ at $R=5$ for the centralized variant versus $15.9$--$20.0$ for the
baselines).

\paragraph{Centralized vs.\ decentralized}
The centralized deployment is the stronger of the two: it achieves lower
average uncertainty and higher minimum and average visit counts at \emph{every}
team size (e.g.\ at $R=3$, uncertainty $0.524$ vs.\ $0.575$, minimum visit
$9.72$ vs.\ $6.47$, and average visit $15.36$ vs.\ $12.92$). This is expected,
as the centralized mapper assimilates every robot's measurements into a single
belief, giving the shared policy a more complete and less noisy map to act on.
The decentralized variant trades some of this performance for looser
coupling (local beliefs fused every $3$ steps); on the reference belief the gap is modest and
\emph{narrows as the team grows} (average
uncertainty gap $0.061\!\rightarrow\!0.051\!\rightarrow\!0.033$ for
$R=2,3,5$), whereas on the robots' own fused beliefs uncertainty barely
decreases with team size ($0.693$, $0.660$, $0.655$: below AUCTION at $R=2$, on
par at $R=3$, above at $R=5$). We attribute this narrowing to the action diversity that per-robot
beliefs introduce: robots conditioned on slightly different local maps are less
prone to collapsing onto the same high-uncertainty region, which becomes
increasingly valuable as more robots share the workspace---indeed, without the
explicit non-collision rule the decentralized variant surpasses the centralized
one at $R=5$ (Table~\ref{tab:nc}: $0.498$ vs.\ $0.571$).

\paragraph{Classical baselines}
AUCTION and COVERAGE exceed the decentralized variant on worst-case fairness at
the smaller team sizes (minimum visit $4.43$/$4.77$ vs.\ $3.53$ at $R=2$), where
an explicit global assignment ensures every target is eventually visited,
while the decentralized variant overtakes them once implicit coordination has
more robots to work with ($13.88$ vs.\ $10.73$/$9.40$ at $R=5$). However, neither
approaches the centralized deployment on any metric, and both leave
substantially higher residual uncertainty than either PROPOSED variant on the reference belief.
The retrained COMPASS performs on par with the classical baselines. The centralized deployment is the
only method that is simultaneously strong on all three metrics, achieving both
low residual uncertainty and balanced, high target coverage.

\subsection{Effect of the explicit non-collision constraint}
\label{subsec:nc-ablation}

The \emph{non-collision} (NC) constraint is an execution-time, conflict-free action-selection rule applied on top of the shared policy:
without NC, each robot greedily takes its most probable action, so two robots may select the \emph{same} next node or edge;
with NC, robots may not select the same node or a node occupied by another robot.

\begin{table}[t]
\centering
\caption{Effect of Non-Collision Constraints ($N=6$)}
\label{tab:nc}
\scriptsize
\setlength{\tabcolsep}{4pt}
\begin{tabular}{l c c c c}
\hline
Variant & $R$ & Unc.$\downarrow$ & Min.Visit$\uparrow$ & Avg.Visit$\uparrow$\\
\hline
\multirow{3}{*}{Cent.}
 & 2 & 0.635 / \textbf{0.582} & 4.50 / \textbf{6.16} & \textbf{11.22} / 10.15 \\
 & 3 & 0.599 / \textbf{0.524} & 8.06 / \textbf{9.72} & \textbf{15.70} / 15.36 \\
 & 5 & 0.571 / \textbf{0.457} & 14.28 / \textbf{17.59} & \textbf{26.80} / 24.24 \\
\hline
\multirow{3}{*}{Decent.}
 & 2 & 0.654 / \textbf{0.643} & 3.31 / \textbf{3.53} & 9.20 / \textbf{9.75} \\
 & 3 & \textbf{0.574} / 0.575 & 6.44 / \textbf{6.47} & \textbf{13.61} / 12.92 \\
 & 5 & 0.498 / \textbf{0.490} & 11.56 / \textbf{13.88} & 21.41 / \textbf{22.20} \\
\hline
\end{tabular}
\\
\footnotesize{Cells: NC-off / NC-on; \textbf{bold} = better per (Variant,$R$) pair.}
\end{table}

NC yields a consistent and sizable improvement for the centralized deployment:
average uncertainty drops at every team size (e.g.\ $0.571\!\rightarrow\!0.457$
at $R=5$), the worst-case target receives more observations
(minimum visit $14.28\!\rightarrow\!17.59$), and the benefit
\emph{grows with team size}. \emph{Average} visits decrease
($26.80\!\rightarrow\!24.24$) because NC redirects redundant co-located
observations to neglected targets. Without NC, robots sharing one belief and
one policy compute nearly identical action preferences and converge onto the
same node (\emph{policy-symmetry collapse}); NC breaks this symmetry.
For the decentralized deployment NC has only a marginal effect (essentially
unchanged at $R=2,3$; a clear improvement only at $R=5$), because
differing per-robot beliefs already diversify actions (\emph{implicit} coordination).

\subsection{Component ablations and scaling}
\label{subsec:ablations}

We ablate each component individually (NC-on, $N=6$, $32$
episodes; $95\%$ CIs $\leq 0.012$ for uncertainty and $\leq 1.2$ for visits).

\paragraph{Belief fusion}
At $R=3$ (decentralized), the reference-belief metric is insensitive to the
fusion weight and exchange rate, but the robots' \emph{own} fused beliefs
degrade monotonically as exchange becomes sparser: own-belief uncertainty at
the targets is $0.658$ exchanging every step, $0.663$ at the default
every-$3$-steps, $0.686$ at every $10$, and $0.756$ with no exchange---a
$14\%$ inflation. The default thus retains essentially the every-step value at
one third of the messages. The fixed $\omega=0.5$ matches a per-exchange
trace-minimizing line search ($0.661$ vs.\ $0.661$).

\paragraph{GP-consistent process update}
Disabling it ($\alpha=1$, no uncertainty inflation) makes the belief
overconfident and removes the incentive to revisit: minimum visit count
collapses from $9.72$ to $3.32$ (centralized) and from $6.47$ to $1.72$
(decentralized) at $R=3$. The temporal model is what sustains persistent
revisiting of moving targets.

\begin{table}[t]
\centering
\caption{Representation ablations (NC-on); cells: first/second variant, \textbf{bold} = better}
\label{tab:repr-ablation}
\scriptsize
\setlength{\tabcolsep}{4pt}
\begin{tabular}{l c c c c}
\hline
Config & $R$ & Unc.$\downarrow$ & Min.Visit$\uparrow$ & Avg.Visit$\uparrow$ \\
\hline
\multirow{3}{*}{\shortstack[l]{Mapper (Cent.):\\diagonal vs.\\full covariance}}
 & 2 & \textbf{0.582} / 0.606 & 6.16 / \textbf{6.63} & 10.15 / \textbf{11.65} \\
 & 3 & \textbf{0.524} / 0.547 & 9.72 / \textbf{10.41} & 15.36 / \textbf{16.43} \\
 & 5 & \textbf{0.457} / 0.480 & 17.59 / \textbf{18.13} & 24.24 / \textbf{25.74} \\
\hline
\multirow{3}{*}{\shortstack[l]{AUCTION planner:\\per-target GPs vs.\\grid belief}}
 & 2 & 0.744 / \textbf{0.607} & 4.43 / \textbf{5.81} & 8.33 / \textbf{9.72} \\
 & 3 & 0.667 / \textbf{0.558} & 7.70 / \textbf{9.16} & 11.76 / \textbf{13.89} \\
 & 5 & 0.566 / \textbf{0.519} & 10.73 / \textbf{14.94} & 15.91 / \textbf{21.52} \\
\hline
\end{tabular}
\end{table}

\paragraph{Covariance representation and planner}
Table~\ref{tab:repr-ablation} (top) evaluates the centralized deployment with
an exact dense-covariance mapper differing from ours only in covariance
storage: it yields slightly better raw throughput (up to $+15\%$ average
visits at $R=2$, shrinking with team size), and in open-loop replays the
diagonal posterior stays within $1.2\%$ of the exact variance---a few percent
of performance in exchange for $4\times$ faster updates and a $512\times$
smaller belief state ($16$\,KB vs.\ $8$\,MB), the payload that makes the
decentralized exchange of Sec.~\ref{subsec:ci} practical.
Table~\ref{tab:repr-ablation} (bottom) runs the \emph{same} AUCTION planner on
both belief representations: the grid GP--Kalman belief alone improves minimum
visits by $19$--$39\%$ and average visits by $17$--$35\%$, and accounts for
$85\%$, $76\%$, and $43\%$ of the centralized deployment's uncertainty gap to
AUCTION at $R=2,3,5$; the centralized policy contributes the remainder and
scales better (minimum visit $14.94\!\rightarrow\!17.59$ at $R=5$). The
decentralized policy, in contrast, does not beat AUCTION on the grid belief in
uncertainty at $R=2,3$ ($0.643$ vs.\ $0.607$; $0.575$ vs.\ $0.558$) nor in
minimum visits at any $R$, although it uses neither a global assignment nor the
centralized belief.
Replaying identical COVERAGE trajectories, hence identical measurement streams,
through both representations with their native sensing models, the grid belief
reports $13\%$, $11\%$, and $6.5\%$ lower uncertainty than per-target GPs at
$R=2,3,5$.

\paragraph{Zero-shot team-size scaling}
Evaluating the same policies far beyond the training team sizes
($R\in\{8,10,15\}$) shows no degradation: uncertainty improves monotonically
($0.397/0.375/0.336$ centralized, $0.430/0.406/0.355$ decentralized, reference belief) and
minimum visits scale with team size ($29.3/38.3/57.4$ centralized), while the
decentralized variant exchanges only fixed-size $16$\,KB packets every $3$ steps.

%% file: experiment.tex
\section{Real-world Experiment}
\label{sec:experiment}

We deploy the proposed approach on a team of two Ascent Aerosystems Spirit
coaxial UAVs, each equipped with a Gremsy ViO gimbal camera
and an NVIDIA
Orin~NX onboard compute board. The two robots communicate over a wireless mesh
network, exchanging compact belief summaries as ROS messages exactly as in the
decentralized simulation deployment; no raw measurements are broadcast. The
team collectively searches for humans in an outdoor $90\times80$~m grass field.
Each robot flies at a fixed altitude of $9$~m with the gimbal camera pitched
$40^\circ$ downward, and human detection is performed onboard with a YOLOv11
detector.

We deploy the \emph{same} decentralized policy as in the simulation benchmark
without fine-tuning; the deployment differs from the trained setting in its
grid ($24\times42$ cells over the field), camera pitch ($40^\circ$ vs.\
$55^\circ$), and exchange period ($5$~s). The targets are two people wandering within a small area. A human detection
produces a positive measurement at the grid cell of the person's estimated
ground location (bounding-box center projected onto the ground plane), and the
non-collision constraint is enabled for safe separation. The policy runs
onboard the Orin~NX with sub-$100$~ms inference; the selected node becomes a
goal point from which a smooth trajectory is sent to the ArduPilot controller.

\label{subsec:exp-results}
The full run is provided in the supplementary video. The team first moves
deep into the unobserved region to explore; once a human is detected, the
policy balances re-observing the person against monitoring regions whose
uncertainty grows over time, the exploration--tracking trade-off observed in
simulation.

We quantify the run from the onboard logs of both vehicles. Over the
$9.8$-minute mission the detectors ran at $10$~Hz and the team first detected a
person after $100$~s. Detected
people were re-observed with a median gap of $12$~s, i.e.\ the policy interleaved re-observation with continued exploration
rather than loitering over a known target. Mean posterior uncertainty over the
mapped area fell by $26\%$ and $25\%$ for the two robots, reaching its minimum
near $t\!\approx\!350$~s and then rising slightly as temporal inflation in
unvisited regions came into balance with new observations.
Each robot broadcast its belief summary (mean and variance over its
$24\times42$ grid) as one $2.1$~KB message every $5$~s, i.e.\ $0.41$~KB/s per
robot and ${\approx}250$~KB for the mission. The exchange was demonstrably useful: the first
detection by one robot appeared in its teammate's fused belief 
$65$~s before that teammate saw the person with its own camera.
This single run, without a baseline or simulated counterpart, is a
feasibility demonstration: both the learned policy and the map representation
transferred to real-world hardware without additional fine-tuning.